\documentclass[sigconf]{acmart}

\AtBeginDocument{%
  \providecommand\BibTeX{{%
    \normalfont B\kern-0.5em{\scshape i\kern-0.25em b}\kern-0.8em\TeX}}}

\copyrightyear{2022} 
\acmYear{2022} 
\setcopyright{acmcopyright}\acmConference[MM '22]{Proceedings of the 30th ACM International Conference on Multimedia}{October 10--14, 2022}{Lisboa, Portugal}
\acmBooktitle{Proceedings of the 30th ACM International Conference on Multimedia (MM '22), October 10--14, 2022, Lisboa, Portugal}
\acmPrice{15.00}
\acmDOI{10.1145/3503161.3548052}
\acmISBN{978-1-4503-9203-7/22/10}

\acmSubmissionID{1264}

\usepackage{xcolor}
\usepackage{bbding} 
\usepackage{algorithm}

\usepackage{algorithm}  
\usepackage{algpseudocode}  
\usepackage{bbding}
\makeatletter
\def\author@bx@sep{0pc}
\makeatother
\begin{document}

\title{TGDM: Target Guided Dynamic Mixup for Cross-Domain Few-Shot Learning}

% \author{Linhai Zhuo$^1$, Yuqian Fu$^1$, Jingjing Chen$^1$, Yixin Cao$^2$, Yu-Gang Jiang$^1$}
% \affiliation{%
 % \institution{$^1$Shanghai Key Lab of Intelligent Information Processing, School of Computer Science, Fudan University}
 % \streetaddress{$\{$19110240043, fuyq20, cjj, ygj@fudan.edu.cn}}

%% author
% \author{Linhai Zhuo}
% \author{Yuqian Fu}
% \affiliation{$\{$lhzhuo19, fuyq20$\}$@fudan.edu.cn}
% \affiliation{Shanghai Key Lab of Intell. Info. Processing, School of CS, Fudan University \\ 
% }
% \settopmatter{authorsperrow=2}
\author{Linhai Zhuo 
\country{}}
\email{lhzhuo19@fudan.edu.cn}
\affiliation{Shanghai Key Lab of Intell. Info. Processing, School of CS, Fudan University \\ 
\country{}
}

\author{Yuqian Fu
\country{}}
\email{fuyq20@fudan.edu.cn}
\affiliation{Shanghai Key Lab of Intell. Info. Processing, School of CS, Fudan University \\ 
\country{}
}

% \author[]{}

\settopmatter{authorsperrow=3}
\author{Jingjing Chen$^\#$
\country{}}
\email{chenjingjing@fudan.edu.cn}
\affiliation{Shanghai Key Lab of Intell. Info. Processing, School of CS, Fudan University \\ 
\country{}
}

\author{Yixin Cao
\country{}}
\email{caoyixin2011@gmail.com}
\affiliation{Singapore Management University \\
\country{}
}

\author{Yu-Gang Jiang
\country{}}
\email{ygj@fudan.edu.cn}
\affiliation{Shanghai Key Lab of Intell. Info. Processing, School of CS, Fudan University \\ 
\country{}
}

% \author[L. Zhuo, Y. Fu, J. Chen, Y. C, Y.-G. Jiang]{Linhai Zhuo$^{1, 2}$, Yuqian Fu$^{1,  2}$, Jingjing Chen$^{1, 2\#}$, Yixin Cao$^3$, Yu-Gang Jiang$^{1, 2}$}
% \affiliation{$^1$Shanghai Key Lab of Intell. Info. Processing, School of CS, Fudan University \\ 
% $^2$ Shanghai Collaborative Innovation Center on Intelligent Visual Computing \\ $^3$Singapore Management University \\
% \country{}
% \country{China}
% }
% \affiliation{$^2$Singapore Management University
% %\country{China}
% }
% \affiliation{\{lhzhuo19, fuyq20, chenjingjing, ygj\}@fudan.edu.cn
% \country{caoyixin2011@gmail.com}}
\thanks{$\#$ indicates corresponding author}

%% short authors
% \renewcommand{\shortauthors}{Zhuo, Fu, Chen and Cao, et al.}
\renewcommand{\shortauthors}{Linhai Zhuo et al.}

\begin{abstract}
Given sufficient training data on the source domain, cross-domain few-shot learning (CD-FSL) aims at recognizing new classes with a small number of labeled examples on the target domain. The key to addressing CD-FSL is to narrow the domain gap and transferring knowledge of a network trained on the source domain to the target domain. To help knowledge transfer, this paper introduces an intermediate domain generated by mixing images in the source and the target domain. Specifically, to generate the optimal intermediate domain for different target data, we propose a novel target guided dynamic mixup (TGDM) framework that leverages the target data to guide the generation of mixed images via dynamic mixup. The proposed TGDM framework contains a Mixup-3T network for learning classifiers and a dynamic ratio generation network (DRGN) for learning the optimal mix ratio. To better transfer the knowledge, the proposed Mixup-3T network contains three branches with shared parameters for classifying classes in the source domain, target domain, and intermediate domain. To generate the optimal intermediate domain, the DRGN learns to generate an optimal mix ratio according to the performance on auxiliary target data. Then, the whole TGDM framework is trained via bi-level meta-learning so that TGDM can rectify itself to achieve optimal performance on target data. Extensive experimental results on several benchmark datasets verify the effectiveness of our method.

\end{abstract}





\maketitle

\begin{figure}[t!]
  \centering
  \includegraphics[width=0.8\linewidth]{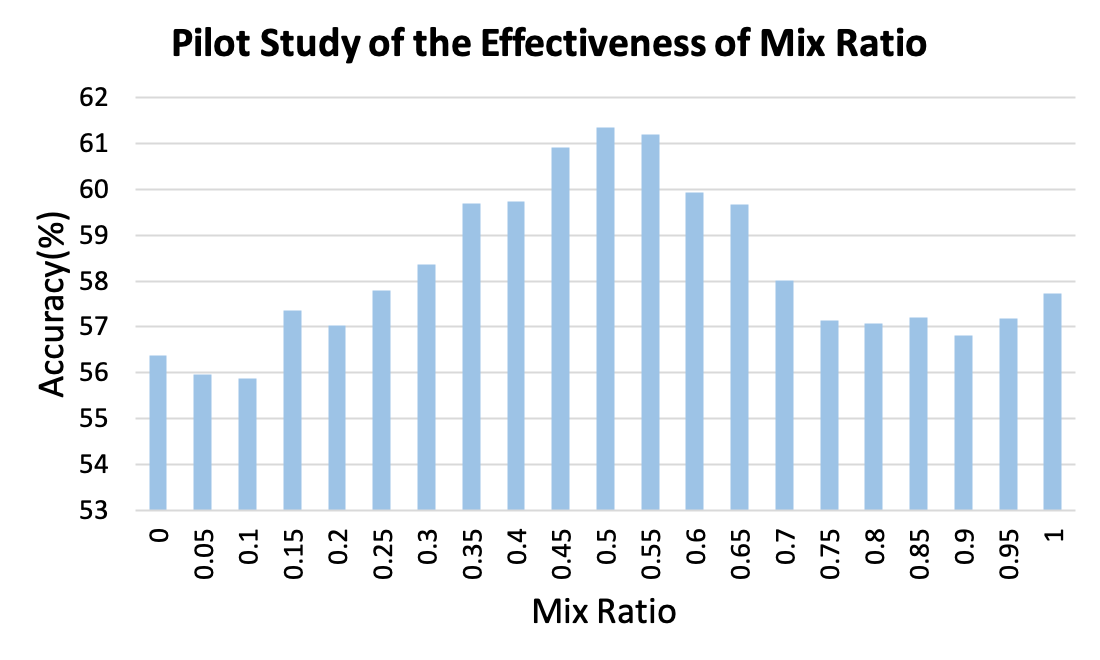}
  \vspace{-0.15in}
  \caption{A pilot study on mixing source and target data with various ratios. We use the proposed network Mixup-3T as the base model.
  %in CD-FSL dataset\yixin{give dataset name and cite}.
  The Mini-Imagenet \cite{ravi2016optimization} and the Places \cite{zhou2017places} are used as the source and target datasets, respectively.}
  \label{fig:pilot}
  \vspace{-0.2in}
\end{figure}

\section{Introduction}

Few-shot learning (FSL) aims to recognize unseen classes with only a few labeled data. This not only helps to alleviate the heavy burden of manual annotations, but also encourages the models to be consistent with human cognitive processes --- we can easily learn new animals or car brands from a couple of examples. FSL will fundamentally benefit many tasks, ranging from object detection \cite{chen2021dual} to video classification \cite{cao2020few}.
Typically, these tasks~\cite{prototypical, sung2018learning, vinyals2016matching, gnn} assume two types of classes: base classes with sufficient data for training, and novel classes with very few data for testing, where base classes and novel classes have no overlap but follow the same data distribution for transfer learning. Clearly, such a distribution assumption is too strong in realistic. Most base and novel classes are actually from different domains in practice. How to mitigate the domain gap for cross-domain few-shot learning (CD-FSL) has attracted increasing attention recently~\cite{tseng2020cross,FWT}.

Building on top of FSL methods, the key challenge of CD-FSL is to improve the model's generalization ability. There are mainly two groups of methods. The first group has no access to the data in target domain and relies on abstracting more discriminative features via adversarial training or disentangle learning. For instance, \citet{wang2021} introduce adversarial task augmentation to improve the robustness of the inductive bias across domains. \citet{fu2022wave} decompose the low-frequency and high-frequency components of images to span the style distributions of the source domain. 

However, the performances of the above methods are still unsatisfactory.
% Another group of methods tends to introduce target domain data and usually achieves superior performance. 
Thus, to achieve superior performance, some methods tend to introduce target domain data. 
The basic idea is to mitigate the domain gap through data augmentation. Except for source domain, \citet{das2021importance} and \citet{liang2021boosting} further fine-tune their models on unlabeled data in target domain via self- or semi-supervised methods, while paper \cite{metafu} demonstrates the effectiveness of using very few labeled target data. Considering the acceptable cost of limited labeled data in practice, we advocate this direction.

In this paper, we propose to investigate the mixup technique \cite{zhang2017mixup} to efficiently use a small amount of labeled target domain data during training for CD-FSL. Mixup is an easy-to-apply data augmentation method. It conducts linear interpolation between source and target domains of data. Thus the mixed data is intermediate between source and target domains. We name it as an intermediate domain throughout the paper. Clearly, training on the data from this intermediate domain not only reconciles the different data distribution from various domains, but also improves the model generalization ability.

Although many works have demonstrated the effectiveness of mixup in various tasks, focusing on CD-FSL, the severe data imbalance issue brings a great challenge --- how to set the ratio of source domain data to target domain data. Focusing more on the limited labeled data in target domain will fall into an over-fitting problem, and less on auxiliary target data may not be helpful in domain adaption.

To show the great impacts of mix ratio on specific CD-FSL tasks, we have conducted a pilot study by choosing the proposed base network Mixup-3T as the model, and Mini-Imagenet \cite{ravi2016optimization} and Places \cite{zhou2017places} as the datasets. As shown in Fig. \ref{fig:pilot}, we can see a great fluctuation of accuracy along with the varying mix ratios. That is, an optimal mixup strategy helps to achieve good performance. Since the optimal mix ratio could be different on different datasets and tasks, it's tedious to choose an optimal mix ratio manually. Therefore, further investigation on mix ratio is necessary, and a well-designed optimization strategy can be beneficial here.

To address the challenges, we propose a novel \textbf{T}arget \textbf{G}uided \textbf{D}ynamic \textbf{M}ixup (TGDM) framework that controls the intermediate domain during training for CD-FSL. By dynamically choosing a suitable mix ratio, TGDM boosts the performance on novel classes, without the harm on base classes. There are two core components: the classification network Mixup-3T and \textbf{D}ynamic \textbf{R}aito \textbf{G}eneration \textbf{N}etwork (DRGN).
The basic idea lies in guided generating mix ratio and utilize intermediate domain effectively. First, based on current mixed data, we optimize Mixup-3T via a tri-task learning mechanism, involving source, target, and intermediate domain classification tasks. The source and target classification tasks target better performance on specific tasks, and the intermediate classification is to improve the generalization ability.
Second, DRGN learns to produce a target guided mix ratio to guide the generation of intermediate domain data. Specifically, We perform a pseudo backward propagation of Mixup-3T to validate the performance on auxiliary target data, whose loss will be utilized to update DRGN.

% Overall, our contributions are summarized as follows:
% % \vspace{-0.15in}
% \begin{itemize}
% \item[1)] We propose a novel target guided dynamic mixup (TGDM) framework that leverages the target data to dynamically control the mix ratio for better generalization ability in cross-domain few-shot learning.
% \item[2)] We propose a Mixup-3T network that utilizes the dynamic mixed data as the intermediate domain for better transferring knowledge between the  source domain and the target domain. 
% \item[3)] We conduct extensive experiments on several benchmarks, and the experimental results demonstrate the effectiveness of our framework.
% \end{itemize} 

Overall, our contributions are summarized as follows:
1) We propose a novel target guided dynamic mixup (TGDM) framework that leverages the target data to dynamically control the mix ratio for better generalization ability in cross-domain few-shot learning.
2) We propose a Mixup-3T network that utilizes the dynamic mixed data as the intermediate domain for better transferring knowledge between the  source domain and the target domain. 
3) We conduct extensive experiments on several benchmarks, and the experimental results demonstrate the effectiveness of our framework.

\section{Related Work}
 %We illustrate the most related from two perspectives: tasks (FSL and CD-FSL) and models (Dynamic Mixup and Meta-Learning). 
%\yixin{the relationship between your model and the following four sections is not clear. you may give a overall para first. Also, you have extra space and the related work is a bit less. }
\subsection{Few-Shot Learning}
Few-shot learning aims at learning new concepts %\yixin{any citation? I am not sure if learning new concept is clear here.} 
with very few samples. Many efforts have been made in this field. These methods are mainly divided into three categories: model initialization \cite{finn2017model, rusu2018meta}, metric-learning \cite{gnn, chen2019image, sung2018learning, vinyals2016matching} and data augmentation \cite{chen2019image, li2020adversarial, fu2019embodied, fu2020depth}. More recently, \citet{zhou2020} apply Similarity Ratio to weight the importance of base classes and thus select the optimal ones. \citet{ji2022information} propose Modal-Alternating Propagation Network to rectify visual features with semantic class attributes. \citet{yan2021} adopt bi-level meta-learning optimization framework to select samples. These methods obtain training and testing images from the same domain. In this paper, we stick to metric learning and bi-level meta-learning but formulate them under the cross-domain scenario with few labeled data.

\subsection{Cross-Domain Few-Shot Learning}
Cross-domain few-shot learning (CD-FSL) aims to perform few-shot classification under the setting where the training and testing data are from different domains. This task is formally defined and proposed by \cite{FWT}. Then, more benchmarks are proposed by \cite{guo2020}. According to whether using target dataset during the training phase \cite{yao2021, startup, liu2021, map, metafu, sun2021explanation} or not \cite{liang2021boosting, FWT, fu2022wave}, CD-FSL methods can be divided into two groups. For training without target data, \citet{wang2021} apply adversarial training to augment data and improve the robustness of the inductive bias. \citet{fu2022wave} believe that the style contains domain specific information. So they transfer styles between two training episodes and apply self-supervised learning to make network ignore the transformation of style. Generally, because of lacking target data, the performances of this kind of methods are lower than that use target data in model training. As a result, some researchers finetune their models on target support set. For example, \citet{liang2021boosting} propose NSAE to enhance feature with noise and \citet{das2021importance} apply contrastive loss. Both of these two works eventually finetune their models with support data in target domains. \cite{yao2021, startup, liu2021} further introduce unlabeled data and turn to additional self-supervised learning tasks on unlabeled target data. \citet{map} integrate several SOTA modules and adaptively chooses the combination of these modules in specific task. Generally, the above mentioned methods are time consuming because they need to finetune network on each test set or require massive unlabeled target data to guarantee the performance of self/semi-supervised methods. Therefore, in this work, we follow the setting proposed in \cite{metafu}, where only a few labeled target data can be used in the FSL, and further explore the dynamic mixup algorithm in this setting. 

 %Besides, \citet{metafu} do detailed pilot study and suggest using very few labeled target data when training. Then, they propose Meta-FDMixup to make use of these data. In this paper, we stick to the CD-FSL with few labeled target data setting \cite{metafu} and further explore the dynamic mixup algorithm in this setting.

\subsection{Dynamic Mixup}
%{\textcolor{red}{T survey dynamic mixup: 1) Mixupxxx; 2) Baisc inculding xxx; 3) The most related work to us is dynamic mixup. (key part)}}
Mixup \cite{zhang2017mixup} is a simple but effective data augmentation method that linearly mixes the two input data according to a random mix ratio. More recently, several variants of mixup are proposed \cite{verma2019manifold, yun2019cutmix, hendrycks2019augmix, kim2020puzzle, TangLPT20, gao2022dynamic}. These methods obtain the mix rate randomly from a certain distribution. Apart from these methods, there are few works that dynamically change the mix ratio during their tasks. \cite{bunk2021, DAT} use learnable mix ratio to enhance the robustness of the model in adversarial training. \citet{guo2019} utilize re-parameterization to learn the prior distribution of the mix ratio. \cite{wu2022, yamaguchi2021} use several predefined ratios to learn model progressively. \citet{mai2021metamixup} are the most related one. They use bi-level meta-learning to learn an optimal mix ratio, but they focus on the general classification task and their network architecture is different from ours. In this paper, we target on CD-FSL problem and manage to control the generation of intermediate domain guided by target data and current model state via dynamic mixup. As far as we are known, we are the first to introduce dynamic mixup under CD-FSL setting.

\subsection{Meta-Learning}
Meta-learning algorithms, also known as learning to learn, aim at enhancing the performance of the network in the new task and are widely used in few-shot learning. The meta-learning manages to optimize the parameters of the network through all the tasks, instead of a certain batch of data. \citet{finn2017model} propose to achieve optimal initial parameters for the new task. \citet{mishra2017simple} propose SNAIL, which consists of a class of simple and generic meta-learner architectures. Specifically, among them, \cite{shu2019meta, ren2018} propose bi-level optimization strategy to reweight the training data. In our paper, we focus on CD-FSL with few labeled target data setting and adopt bi-level meta-learning, so that the TGDM framework can generate target guided mixed data. 

% \caption{Pipeline of the proposed TGDM framework. In the first level, given an training episode and target validation loss, the Mixup-3T is the classification network and DRGN is to produce guided mix ratio for better optimizing Mixup-3T. Furthermore, a bi-level meta-learning strategy is adopted in our framework to take one gradient step and validate on auxiliary target data to find the optimal gradient direction for the target domain.  ``BP" is the abbreviation of ``Backward Propagation".} 

\section{Method}

\begin{figure*}[h]
  \centering
  \includegraphics[width=0.80\textwidth]{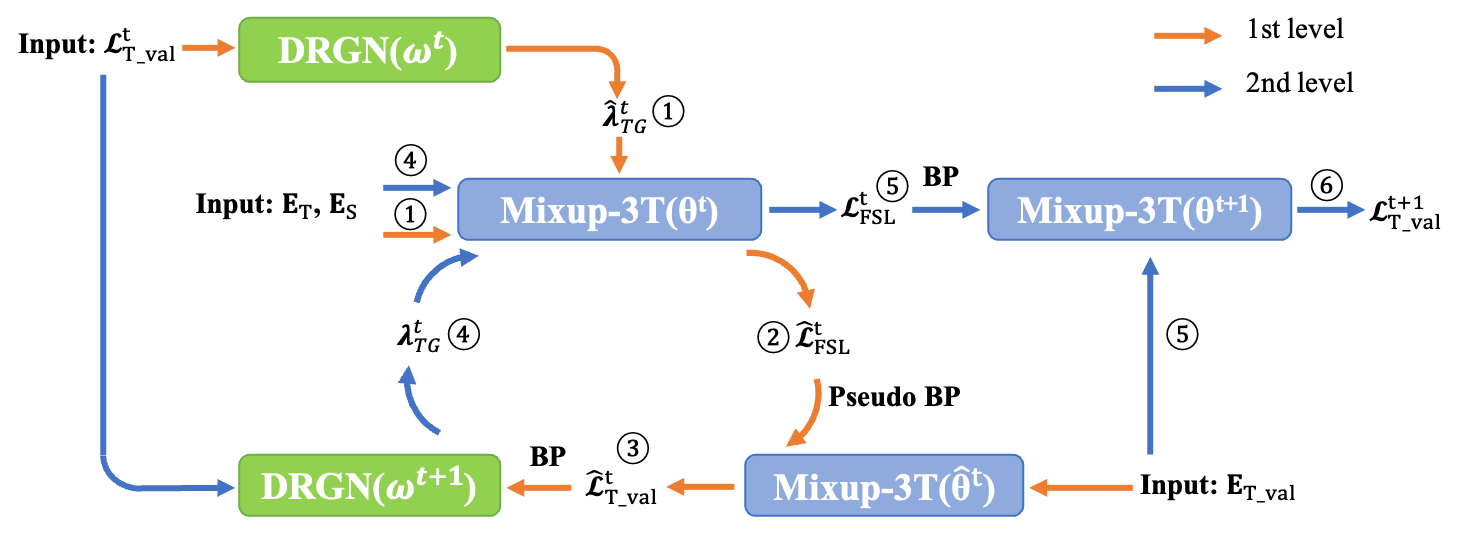}
  \caption{Pipeline of the proposed TGDM framework. In the first level, the Mixup-3T conducts pseudo backward propagation, given a training episode and guided mix ratio generated by DRGN. Then, the updated Mixup-3T is validated on auxiliary target data. Next, the validation loss is utilized to update DRGN. In the second level, the same training episode together with the regenerated guided mix ratio are given to Mixup-3T, for the purpose of conducting vanilla backward propagation. "BP" is the abbreviation of "Backward Propagation".} 
  %\yixin{smaller the figure.}}
  \label{fig:framework}
\end{figure*}

\subsection{Preliminaries}
\textbf{Problem Formulation.} Given source domain dataset ${D_S}$ and the target domain dataset ${D_T}$. 
Three sub-datasets including source base data $D_{S_b} = \{ {I_j},{y_j}\}_j^{{M_{S}}}$, target novel data $D_{T_n}$ and auxiliary target data $D_{aux} = \{ {I_j},{y_j}\} _j^{{M_{aux}}}$ are further obtained from $D_S$ and ${D_T}$. 
Specifically, the classes of $D_{S_b}$ are randomly sampled from $D_S$, while $D_{T_n}$ and $D_{aux}$ are sampled from ${D_T}$. 
%The classes of these three sub-datasets are disjoint from each other. It is worth to note that the number of data $M_{S} >> M_{aux}$.
It is worth to note that the class sets of these three sub-datasets are disjoint from each other and we have $M_{S} >> M_{aux}$.
%In this setting, the network is firstly trained on source base dataset $D_{S_b}$ and auxiliary target dataset $D_{aux}$. Then the network is tested via N-way k-shot classification tasks on the target novel dataset $D_{T_n}$. 
In this setting, the network is trained on $D_{S_b}$ and $D_{aux}$, and tested on $D_{T_n}$. 
\par
% Typically, the N-way k-shot classification tasks are performed to learn meta-knowledge.
%Concretely, $N$ classes with $k$ support images and some queries are obtained from $D_T$. 
% Concretely, $N$ classes with $k$ support images and some queries are sampled as a meta-task.
% The network is supposed to assign labels for these query images assisted with support images.\par
% \noindent\textbf{N-way k-shot Problem.} We sample training data in meta-learning way to mimic the testing process in N-way k-shot classification. During training, we sample an episode consisting of N classes and $k+n_q$ images for each class, where $n_q$ is the number of testing images. Furthermore, the labels of these N classes are re-written as $[0, 1..., N-1]$. 
\noindent\textbf{N-way k-shot Problem.} Typically, the N-way k-shot classification tasks are performed to learn meta-knowledge. By meta-training and meta-testing the model in the same form of N-way k-shot meta-tasks (episodes), it learns to mimic the testing process thus transferring the knowledge well. Concretely, for each meta task, $N$ classes with $k$ support images and $n_q$ queries are sampled. The network is supposed to assign labels for these query images assisted with the support images. Furthermore, the labels of these N classes are re-written as $[0, 1..., N-1]$.

\subsection{Overview}
The TGDM framework mainly consists of two modules. One is the main classification network Mixup-3T $f_{M3}(I|\theta)$. And the other is the dynamic ratio generation network DRGN $g_{G}(z|\omega)$. The $\theta$ and $\omega$ are the parameters of these two networks while $I$ and $z$ are inputs. 
1) Given an image, Mixup-3T aims at predicting the correct class from a list of predefined classes, where the image can be sampled from either source or target domain. 
During training, Mixup-3T will mix both base- and novel-domain data as intermediate domain data according to given ratios dynamically. For validation and inference, it takes only novel-domain images.
2) DRGN aims at updating the mix ratio to better optimize Mixup-3T network. Thus, it validates the current performance of Mixup-3T and leverages the validation loss $L_{TG}$ to estimate a optimal ratio $\lambda_{TG}$ for next iteration. 
%The DRGN takes the target validation loss $L_{TG}$ as input and generates the guided mix ratio $\lambda_{TG}$, which is further sent to Mixup-3T.
Given the above two modules, the overall TGDM framework is trained via a bi-level meta-learning strategy. In the following paragraph, we first introduce the two proposed modules, $f_{M3}$ and $g_G$, then illustrate the training and inference processes of TGDM framework, which is illustrated in Fig. \ref{fig:framework}. 

\begin{figure}[h]
  \centering
  \includegraphics[width=0.8\linewidth]{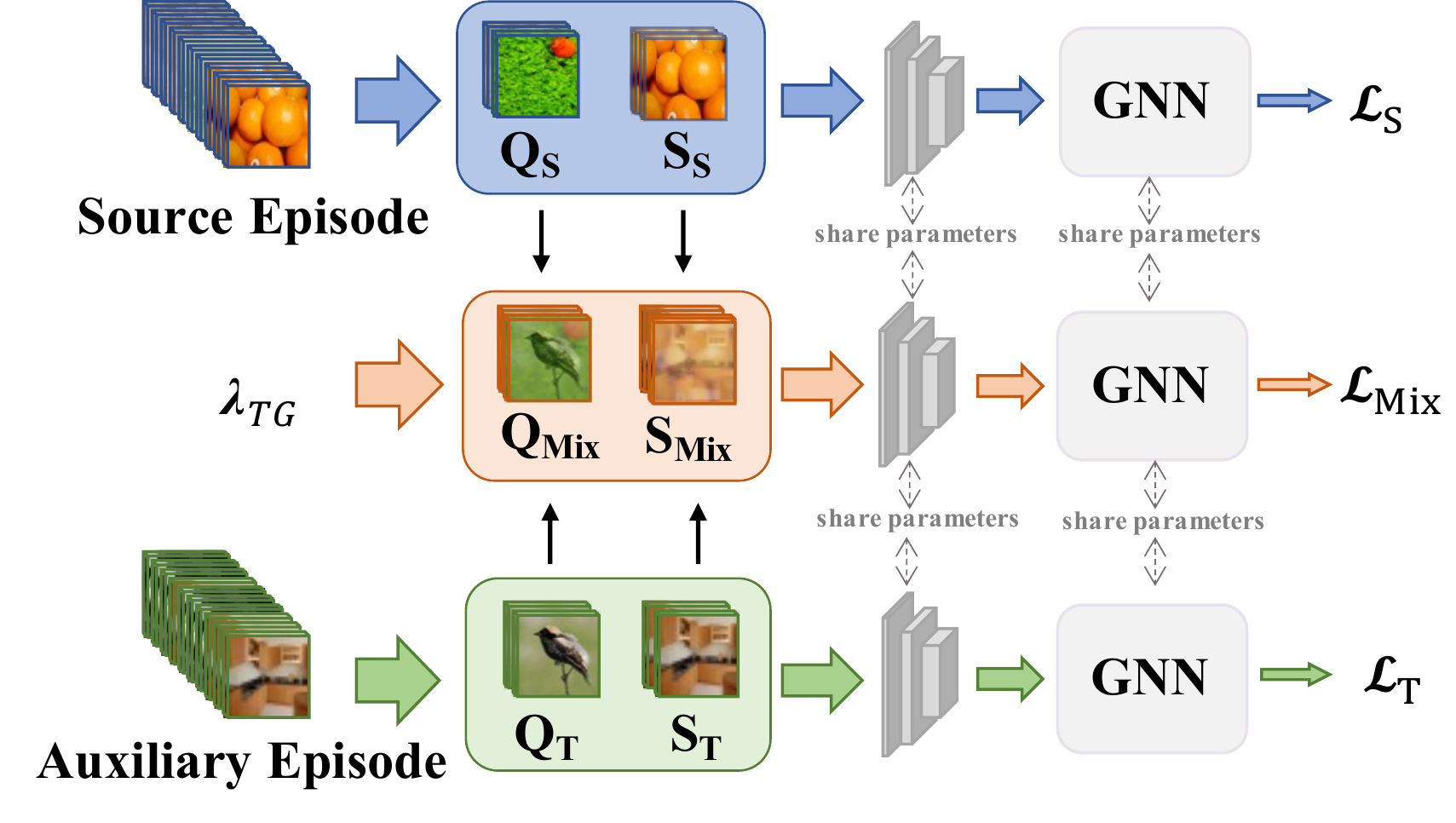}
  \caption{Illustration of our Mixup-3T network. The target guided mix ratio is utilized to guide the mixture of the two episode from source and auxiliary target data. Tri-task learning mechanism including source, target, and intermediate domain classification is adopted to help knowledge transfer.}
  \vspace{-0.2in}
  \label{fig:m3t}
\end{figure}
\subsection{Modules}
\textbf{Mixup-3T.} The detailed illustration of Mixup-3T network is shown in Fig. \ref{fig:m3t}. The feature extractor backbone, and the GNN \cite{gnn} classifier are defined as $f(x)$ and $g_{cls}(F)$, respectively. $x$ and $F$ stand for the inputs images and vector representation of images respectively. Specifically, $f$ can be replaced by any image feature extractor. The Mixup-3T aims at generating intermediate domain data guided by $\lambda_{TG}$ and makes full use of source, target, and intermediate domain data to transfer knowledge. Concretely, the target guided mix ratio $\lambda_{TG}$ together with images from the different domains are taken as input. Then the mixed data are generated according to $\lambda_{TG}$. The feature extractor $f$ extracts a vector representation $F$ from input image. After that GNN classifier receives the vector representation $F$ of both query $Q$ and support $S$ data. Then, the GNN classifier produces prediction logit for each input query.\par
% During training, the Mixup-3T network takes an episode from source base data $E_{S}$ and an episode from auxiliary target data $E_{T}$ as input. 
During training, the Mixup-3T network takes two episode from source base data $E_{S}$ and auxiliary target data $E_{T}$ as inputs. 
Then, $E_{S}$ is further divided into query set $Q_S$ and support set $S_S$. Likewise, $E_{T}$ is divided into $Q_T$ and $S_T$. The according label of $Q_S$ and $Q_T$ is $y_{QS}$ and $y_{QT}$. Then, images from the source and target domain are mixed according to Eq. \ref{eq:mixq}, thus producing $Q_{Mix}$ and $S_{Mix}$: 
% \begin{equation}
%     \label{eq:mixq}
%     Q_{Mix} = \lambda_{TG} Q_S + (1-\lambda_{TG})Q_T\\
% \end{equation}
% \begin{equation}
%     \label{eq:mixs}
%     S_{Mix} = \lambda_{TG} S_S + (1-\lambda_{TG})S_T\\
% \end{equation}
\begin{equation}
    \label{eq:mixq}
    Q_{mix} = \lambda_{TG} Q_S + (1-\lambda_{TG})Q_T, \  S_{mix} = \lambda_{TG} S_S + (1-\lambda_{TG})S_T \\
\end{equation}
The mixed data $Q_{Mix}$ and $S_{Mix}$ are supposed to be in the same intermediate domain, for we use the same mix ratio to mix data between $E_T$ and $E_S$. Technically, $[C_0^S, C_1^S, ..., C_N^S]$ are the classes from $E_S$ and $[C_0^T, C_1^T, ..., C_N^T]$ are the classes from $E_T$. We first choose class pairs accordingly, for instance, $C_0^S$ and $C_0^T$, $C_N^S$ and $C_N^T$ etc, then mix the images between the chosen class pairs and total N new mixed classes are generated. Since the classes are mixed pair-wise, we regard the mixed images from the same class pair are of the same class. We further assign labels from $[0, 1..., N-1]$ to these mixed data. Then, the feature extractor $f$ is used to extract $1-D$ feature representation $F_{QS}$, $F_{SS}$, $F_{QT}$, $F_{ST}$, $F_{QM}$ and $F_{SM}$ from $Q_S$, $S_S$, $Q_T$, $S_T$, $Q_{Mix}$ and $S_{Mix}$ accordingly.\par
A tri-task learning mechanism is proposed for Mixup-3T. This tri-task consists of three classification tasks in source, target, and intermediate domains, respectively. The source domain classification task is supposed to maintain the ability to recognize source images, while the target domain classification task aims to achieve good performance in target domain. The intermediate domain classification task classifies images in the intermediate domain. The intermediate domain classification task is designed to utilize data in the intermediate domain to assist the knowledge transfer. These three tasks share the same network architecture. Concretely, GNN classifier received query set and support set of according domain: $F_{QS}$, $F_{SS}$ for source domain classification; $F_{QT}$, $F_{ST}$ for the target domain classification and $F_{QM}$, $F_{SM}$ for intermediate domain classification. Then, $g_{cls}$ utilizes the cross-entropy (CE) algorithm to calculate classification loss.\par 
The losses of the three tasks are defined as $L_S$, $L_T$ and $L_{Mix}$, respectively. And the overall loss $L_{FSL}$ is calculated as, 
\begin{align}
    L_S &= \sum CE(g_{cls}(f(Q_{S}), f(S_{S})), y_{QS}) \\
    L_T&= \sum CE(g_{cls}(f(Q_{T}), f(S_{T})), y_{QT}) \\
    L_{Mix} &= \sum CE(g_{cls}(f(Q_{Mix}), f(S_{Mix})), y_{QM})
\end{align}
\begin{equation}
    \label{eq:lfsl}
    L_{FSL} = \alpha_0 L_S + \alpha_1 L_T + \alpha_2 L_{Mix}\\
\end{equation}
\noindent where $\alpha_0$, $\alpha_1$, $\alpha_2$ are hyper-parameters, $y_{QM}$ is the label of $Q_{Mix}$.\par
During validation and testing, Mixup-3T takes only one episode as input. Concretely, the episode $E_{T\_val}$ from the auxiliary target dataset $D_{aux}$ is used for validation and the episode $E_{T\_test}$ sampled from the target novel dataset $D_{Tn}$ is for testing. Likewise, $E_{T\_val}$ is divided into query set $Q_{T\_val}$ and support set $S_{T\_val}$, and $E_{T\_test}$ is divided into query set $Q_{Test}$ and support set $S_{Test}$. Then, the features are extracted and sent to the GNN classifier to produce logit. For validation, cross-entropy loss is used to calculate the target validation loss $L_{T\_val}$. As for testing, the class of max output logit is regarded as the prediction. The equations for validation and testing are written in Eq. \ref{eq:loss_val} and Eq. \ref{eq:classify}:
\begin{equation}
    \label{eq:loss_val}
    L_{T\_val} = \sum CE(g_{cls}(f(Q_{T\_val}), f(S_{T\_val})), y_{QTV})\\
\end{equation}
\begin{equation}
    \label{eq:classify}
    C = argmax \ \ g_{cls}(f(Q_{Test}), f(S_{Test}))
\end{equation}
% where $S_{T\_val}$ and $S_{T\_val}$ are support target validation set and query target validation set respectively, 
where C is the prediction class, $y_{QTV}$ is the label of $Q_{T\_val}$.\par
\noindent\textbf{DRGN.} The dynamic ratio generation network is made up of three FC layers followed by a sigmoid activation, for the insight that three layers of MLP have the ability to approximate almost any function. To dynamically generate the mix ratio according to the current state of Mixup-3T network, the ratio generation network takes the target validation loss $L_{T\_val}$ of Mixup-3T as input and generates the target guided mix ratio $\lambda_{TG}$, according to Eq. \ref{eq:tgrn}:
\begin{equation}
    \label{eq:tgrn}
    \lambda_{TG} = g_G(L_{T\_val})
\end{equation}
\subsection{Training and Inference}
\noindent\textbf{Training.} We adopt a bi-level meta-learning strategy to train the entire TGDM framework. It consists of two nested optimization levels. Concretely, in the first level, the framework takes a single gradient step and then rectifies DRGN on auxiliary target data to achieve good performance in the target domain. The second level is to use the rectified DRGN to produce guided mix ratio $\lambda_{TG}$ for Mixup-3T training. For better identification, we use the symbol caret ( $\hat{}$ ) to distinguish data of different levels. With this strategy, both modules of Mixup-3T and DRGN will be optimized iteratively. The pseudo-code of this process is illustrated in Algorithm  \ref{algorithem1}.\par
In the first level, we conduct pseudo backward propagation to Mixup-3T and validate the performance of Mixup-3T on auxiliary target data, then, update the parameters of DRGN. Concretely, two episodes $E_S$ and $E_T$ are randomly sampled from $D_{S_b}$ and $D_{aux}$, respectively, while each episode contains $N*(k+n_q)$ images. In order to avoid overfitting problem, we do not reuse $E_{T}$ as validation data, but sample another episode $E_{T\_val}$ of $N*(k+n_q)$ images randomly from $D_{aux}$. Then, the target validation loss $L_{T\_val}^t$ calculated from the last iteration is fed into DRGN to produce the mix ratio $\hat {\lambda} ^t$. As for the first iteration, we calculate $L_{T\_val}^t$ on auxiliary target data according to Eq. \ref{eq:loss_val}. After $\hat {\lambda}_{TG} ^t$ is calculated, two episodes $E_{S}$ and $E_{T}$ together with mix ratio $\hat {\lambda}_{TG}^t$ are fed into Mixup-3T network and the few-shot classification loss $\hat {L}_{FSL}^t$ is calculated according to Eq. \ref{eq:lfsl}. After that, we do pseudo backward propagation by $\hat {L}_{FSL}^t$ and update $\theta^t$ as $\hat{\theta}^t$. Technically, the pseudo backward propagation refers to doing backward propagation and updating parameters on a copy of Mixup-3T network instead of the original one. Then, to validate the performance of this gradient descent step, $E_{T\_val}$ are fed into Mixup-3T and the target validation loss $\hat {L}_{T\_val}^t$ is calculated according to Eq. \ref{eq:loss_val}. Finally, we do backward propagation to update $\omega$ as $\omega^{t+1}$ with $\hat {L}_{T\_val}^t$ for the purpose of utilizing the auxiliary target data to guide the mixup operation.\par
After optimized with the target validation loss $\hat {L}_{T\_val}^t$, DRGN is supposed to regenerate an optimal mix ratio $\lambda_{TG}$ for target domain and current model state. Therefore, in the second loop, we use the regenerated $\lambda_{TG}$ to optimize Mixup-3T. 
%\yixin{what do you mean by consistence here? Do you mean they are the same?} 
Concretely, $L_{T\_val}^t$ is firstly fed into DRGN to regenerate $\lambda_{TG}^t$ by Eq. \ref{eq:tgrn}. Secondly, $E_{T}$, $E_{S}$ together with $\lambda_{TG}^t$ are fed into Mixup-3T network to calculate few-shot classification loss $ L_{FSL}^t$ by Eq. \ref{eq:lfsl}. Then, $L_{FSL}^t$ is backward propagated to optimize $\theta^t$ and get the $\theta^{t+1}$. At last, we validate the performance of Mixup-3T on auxiliary target data $E_{T\_val}$ and the according loss $L^{t+1}_{T\_val}$ is calculated for next iteration by Eq. \ref{eq:loss_val}.\par
\noindent\textbf{Inference.} As for inference, an episode $E_{T\_test}$ of $N*(k+n_q)$ images in N-way k-shot form is randomly sampled from $D_{T_n}$. Then, $E_{T\_test}$ is fed into Mixup-3T to predict labels according to Eq. \ref{eq:classify}.
\begin{algorithm}[htb]
\caption{TGDM framework}
\label{algorithem1}
\begin{algorithmic}[1]
\Require
Source base dataset $D_{S_b}$, auxiliary target data $D_{aux}$, max iteration T
\Ensure
Mixup-3T $f_{M3}$ network's parameter $\theta^T$
\State Initialize the Mixup-3T network parameter $\theta^0$, DRGN parameter $\omega^0$ and t = 0
\State Sample episode $E_{T\_val}$ from $D_{aux}$
\State Calculate target validation loss $L_{T\_val}^0$ by Eq. \ref{eq:loss_val}
\While {t < T}
\State Sample episodes: $E_{S}$ from $D_{S_b}$ and $E_{T}$, $E_{T\_val}$ from $D_{aux}$
\State Calculate target guided mix ratio $\hat{\lambda}^t_{TG}$ by Eq. \ref{eq:tgrn}
\State Input $E_{S}$, $E_{T}$ to $f_{M3}$ and obtain $\hat{L}_{FSL}^t$ by Eq. \ref{eq:lfsl}
\State Update $\hat{\theta}^t$ by pseudo backward propagation with $\hat{L}_{FSL}^t$
\State Input $E_{T\_val}$ to $f_{M3}$ and obtain $\hat{L}_{T\_val}^t$ by Eq. \ref{eq:loss_val}
\State Update $\omega^{t+1}$ by backward propagation with $\hat{L}_{T\_val}^t$
\State Calculate $\lambda^t_{TG}$ by Eq. \ref{eq:tgrn}
\State Input $E_{S}$, $E_{T}$ to $f_{M3}$ and obtain $L_{FSL}^t$ by Eq. \ref{eq:lfsl}
\State Update $\theta^{t+1}$ by backward propagation with $L_{FSL}^t$
\State Use $E_{T\_val}$ as input and obtain $L_{T\_val}^{t+1}$ by Eq. \ref{eq:loss_val}
\EndWhile
% \vspace{-0.1in}
\end{algorithmic}
\end{algorithm}

\section{Experiment}
%\subsection{Dataset}
\noindent\textbf{Datasets:} Mini-Imagenet \cite{ravi2016optimization} is utilized as source dataset $D_S$, and the other four datasets, including CUB \cite{wah2011caltech}, Cars \cite{krause20133d}, Places \cite{zhou2017places}, and Plantae \cite{van2018inaturalist} are regarded as target datasets $D_T$. We follow meta-FDMixup \cite{metafu} and obtain three disjoint sub-datasets: 1) source base dataset $D_{Sb}$, 2) auxiliary target dataset $D_{aux}$, and 3) target novel dataset $D_{Tn}$. 
Specifically, we train our framework on $D_{Sb}$ and $D_{aux}$, and test on $D_{Tn}$.

\noindent\textbf{Implementation Details:}
%A pre-trained ResNet10 \cite{he2016deep} is used as the feature extractor in our framework, the parameters of which are provided by \cite{FWT}. 
A pre-trained ResNet10 \cite{he2016deep} provided by FWT~\cite{FWT} is used as the feature extractor.
The hyper-parameters $\alpha_0$, $\alpha_1$ and $\alpha_2$ in Eq. \ref{eq:lfsl} are assigned with 0.25, 0.25 and 0.5, respectively. We train the whole network for 400 epochs and every epoch has 100 iterations. Therefore, the max-iteration $T$ is set to 40000. The model trained by the last epoch is recorded for testing. Adam is utilized as the optimizer in the training strategy, and the initial learning rates of Mixup-3T and DRGN are 0.001 and 0.0001, respectively. The weight decay of DRGN is set to 1e-5. During testing, we do 5-way k-shot classification on $D_{T_n}$. Concretely, we randomly sample an episode from $D_{T_n}$ and feed it into the Mixup-3T to test the performance. For a certain episode, 5 classes, k support, and 15 query images are contained. This testing operation is repeated 10000 times, and the mean accuracy is reported as the final result.

%\subsection{Competitors and Baselines}
\noindent\textbf{Competitors and Baselines:} 
Several FSL and CD-FSL methods are compared including MatchingNet \cite{vinyals2016matching}, RelationNet \cite{sung2018learning}, GNN \cite{gnn}, ATA \cite{wang2021}, Wave-SAN \cite{fu2022wave} and Meta-FDMixup \cite{metafu}. Except \cite{metafu}, the other competitors do not use auxiliary data. The backbones of these works are unified to ResNet10. Besides, we add the auxiliary target data to the original training data of \cite{sun2021explanation} and \cite{fu2022wave}, re-run these two methods, and mark them as m-LRP and m-WaveSAN. %We use three baselines in our experiment s-base, m-base, and base-M3T. 
Three baselines including s-base, m-base, and base-M3T are used.
The s-base is trained only with source data. 
%The m-base is trained with source data together with auxiliary target data, which is the simplest way to utilize auxiliary target data. 
The m-base is trained with the merged data of the source images and the auxiliary target images, which is the simplest way to utilize the auxiliary target data. 
The other baseline is base-M3T. 
%Base-M3T is to train Mixup-3T without target guided.
%Base-M3T is to train Mixup-3T without guidance from target data.
Base-M3T is to train Mixup-3T with the vanilla mixup which the mix ratio is not guided by the target data.
%Concretely, in base-M3T, we do not utilize DRGN to produce $\lambda_{TG}$ for Mixup-3T network, but randomly choose $\lambda_{TG}$ from Beta(1.0, 1.0) distribution. 
Concretely, in base-M3T, rather than generating $\lambda_{TG}$ by DRGN, the  mix ratio is ramdonly sampled from the Beta distribution.
All the baselines use ResNet10 as the feature extractor and GNN as the classifier.

\begin{table*}[t]
    \centering
    \begin{tabular}{lllllll}
    \hline
    5-way 1-shot &&$D_{aux}$&CUB&Cars&Places&Plantae\\
    \hline
    FSL&MatchingNet\cite{vinyals2016matching}&-&35.89$\pm$0.51&30.77$\pm$0.47&49.86$\pm$0.79&32.70$\pm$0.60\\
       &RelationNet\cite{sung2018learning}&-&42.44$\pm$0.77&29.11$\pm$0.60&48.64$\pm$0.85&33.17$\pm$0.64\\
       &GNN\cite{gnn}&-&45.69$\pm$0.68&31.79$\pm$0.51&53.10$\pm$0.80&35.60$\pm$0.56\\
    \hline
    CD-FSL&ATA\cite{wang2021}&-&45.00$\pm$0.50&33.61$\pm$0.40&53.57$\pm$0.50&34.42$\pm$0.40\\
          &Wave-SAN\cite{fu2022wave} &-&50.25$\pm$0.74&33.55$\pm$0.61&57.75$\pm$0.82&40.71$\pm$0.66\\
          &m-LRP\cite{sun2021explanation} & \checkmark & 59.23$\pm$0.58 & 46.88$\pm$0.53 & 57.92$\pm$0.58 & 49.11$\pm$0.54\\
          &m-WaveSAN\cite{fu2022wave} & \checkmark & 63.59$\pm$0.85 & 50.06 $\pm$0.76 & 59.89$\pm$0.86 & 51.99$\pm$0.81\\
    &meta-FDMixup\cite{metafu}&\checkmark&63.24$\pm$0.82&\textbf{51.31$\pm$0.83}&58.22$\pm$0.82&51.03$\pm$0.81\\
     \hline
    % baseline&s-base\cite{metafu}&-&45.69$\pm$0.68&31.79$\pm$0.51&53.10$\pm$0.80&35.60$\pm$0.65\\
    Baseline&m-base\cite{metafu}&\checkmark&57.65$\pm$0.80&46.03$\pm$0.72&55.70$\pm$0.79&48.25$\pm$0.74\\
    &base-M3T (ours)&\checkmark&63.87$\pm$0.26&49.02$\pm$0.24&59.99$\pm$0.26&51.65$\pm$0.25\\
    \hline
    &TGDM (ours)&\checkmark&\textbf{64.80$\pm$0.26}&50.70$\pm$0.24&\textbf{61.88$\pm$0.26}&\textbf{52.39$\pm$0.25}\\
    \hline
    \hline
    5-way 5-shot&&$D_{aux}$&CUB&Cars&Places&Plantae\\
    \hline
     FSL methods   &MatchingNet\cite{vinyals2016matching}&-&51.37$\pm$0.77&38.99$\pm$0.64&63.16$\pm$0.77&46.53$\pm$0.68\\
       &RelationNet\cite{sung2018learning}&-&57.77$\pm$0.69&37.33$\pm$0.68&63.32$\pm$0.76&44.00$\pm$0.60\\
       &GNN\cite{gnn}&-&62.2$\pm$50.65&44.28$\pm$0.63&70.84$\pm$0.65&52.53$\pm$0.59\\
    \hline
    CD-FSL&ATA\cite{wang2021}&-&66.22$\pm$0.50&49.14$\pm$0.40&75.48$\pm$0.40&52.69$\pm$0.40\\
       &Wave-SAN\cite{fu2022wave} & - &70.31$\pm$0.67&46.11$\pm$0.66&76.88$\pm$0.63&57.72$\pm$0.64\\
       &m-LRP\cite{sun2021explanation} & \checkmark &77.07$\pm$0.44&64.38$\pm$0.48&77.73$\pm$0.45&67.90$\pm$0.47\\
       &m-WaveSAN\cite{fu2022wave} & \checkmark & 82.29$\pm$0.58 & 66.93$\pm$0.71&80.01$\pm$0.60&71.27$\pm$0.70\\
    % &Metamixup\cite{mai2021metamixup}&\checkmark& 79.56$\pm$0.19&65.09$\pm$0.21&75.22$\pm$0.20&65.63$\pm$0.22\\
    &meta-FDMixup\cite{metafu}&\checkmark&79.46$\pm$0.63&66.52$\pm$0.70&78.92$\pm$0.63&69.22$\pm$0.56\\
     \hline
    %   Baseline&s-base\cite{metafu}&-&62.65$\pm$0.65&44.28$\pm$0.63&70.84$\pm$0.65&52.53$\pm$0.58\\
     Baseline&m-base\cite{metafu}&\checkmark&78.08$\pm$0.60&63.27$\pm$0.70&75.90$\pm$0.67&66.69$\pm$0.68\\
      &base-M3T (ours)&\checkmark&82.14$\pm$0.19&67.58$\pm$0.22&81.09$\pm$0.19&69.19$\pm$0.22\\
    \hline
        &TGDM (ours)&\checkmark&\textbf{84.21$\pm$0.18}&\textbf{70.99$\pm$0.21}&\textbf{81.62$\pm$0.19}&\textbf{71.78$\pm$0.22}\\
    \hline 
    \end{tabular}
    \caption{The accuracy(\%) of four target datasets under 5-way 1-shot and 5-way 5-shot. Among all the competitors and baselines, our TGDM framework achieves the best performance in most cases. %except that TGDM is slightly lower than meta-FDMixup in 5-way 1-shot cars setting by 0.61\%
    }
    \vspace{-0.1in}
    \label{tab:main_result}
\end{table*}

\begin{table*}[]
    \centering
    \begin{tabular}{llllll}
    \hline
       Target Dataset &Validation& Mini-ImageNet | CUB & Mini-ImageNet | Cars & Mini-ImageNet | Places & Mini-ImageNet | Plantae\\
    \hline
       s-base\cite{metafu}&-&80.87$\pm$0.56&80.87$\pm$0.56&80.87$\pm$0.56&80.87$\pm$0.56\\
       m-base\cite{metafu}&\checkmark&78.94$\pm$0.58&80.75$\pm$0.55&79.99$\pm$0.58&80.51$\pm$0.55\\
       Meta-FDMixup\cite{metafu}&\checkmark&82.29$\pm$0.57&81.00$\pm$0.58&81.37$\pm$0.56&79.64$\pm$0.59\\
       base-M3T&-&\textbf{83.85$\pm$0.16}&82.87$\pm$0.18&84.51$\pm$0.17&\textbf{81.82$\pm$0.17}\\
       TGDM&-&83.58$\pm$0.17&\textbf{83.33$\pm$0.18}&\textbf{85.80$\pm$0.16}&81.79$\pm$0.18\\
    \hline
    \end{tabular}
    \caption{The accuracy(\%) of Mini-ImageNet under 5-way 5-shot setting. This experiment is conducted to verify that our models can improve target domain performance without the harm on the source domain. The column 'Validation' means whether a method chooses the parameters of the best performance on source validation data.}
    \label{tab:mini}
    %\vspace{-0.05in}
\end{table*}

\subsection{Main Result}
\textbf{Results on four target datasets.} 5-way 5-shot task and 5-way 1-shot tasks are conducted in this experiment. The experimental results are summarized in Table \ref{tab:main_result}. In Table \ref{tab:main_result}, we compare our method with baselines and competitors. 
From Table \ref{tab:main_result}, we have the following observations: First of all, our TGDM framework outperforms the other baselines in most cases. Specifically, our framework achieves 84.21\%, 70.99\%, 81.62\%, and 71.78\% on CUB, Cars, Places, and Plantae under 5-way 5-shot setting. Our method outperforms the SOTA m-WaveSAN by 1.92\%, 4.06\%, 1.61\% and 0.51\%, 
respectively, which indicates the effectiveness of our TGDM framework. It's also verified that utilizing target guided intermediate domain can help knowledge transfer from the source domain to the target domain. As for 5-way 1-shot setting, TGDM performs best in CUB, Places, and Plantae datasets, exceeding SOTA by 1.21\%, 1.99\%, and 0.40\%. For the Cars dataset, our TGDM framework is very slightly lower than Meta-FDMixup by 0.61\%. Given that the disentangle module is the main contribution of \cite{metafu}, we explain this module works effectively, for the Cars dataset is very distinct from Mini-Imagenet. 
%\yixin{not very straightforward. Also, you can say your framework can do disentangle in the future?} 
Secondly, the methods of CD-FSl with few labeled data are higher than other methods, which shows this setting can advance the performance of CD-FSL. Besides, the performances of standard FSL learning methods are much lower than CD-FSL method. It's reasonable that standard FSl learning methods are not designed specifically for knowledge transfer. Thirdly, the base-M3T achieves almost the same performances as the SOTA (Meta-FDMixup and m-WaveSAN), which indicates that the simple randomly chosen mixed data could also benefit the knowledge transfer.\par

\noindent\textbf{Results on Mini-Imagenet.} we also test the performance of our models under 5-way 5-shot setting on Mini-Imagenet to verify that our framework can improve target domain performance without the harm on the source domain. We obtain the parameters of our model from the last iteration, while Meta-FDMixup and m-base choose the parameters of the best performance on source validation data.
%\yixin{if this uses different settings, you can give a separate section.} The experiment results are shown in Table.\ref{tab:mini}. 
From Table \ref{tab:mini}, it can be observed that: 1) Our models (TGDM and Mixup-3T) still achieve the best performance without using the source validation dataset. Since we use a tri-task learning mechanism in our models, it further indicates that the tri-task learning mechanism can help models to achieve good performance on both source and target datasets.
2) Generally, mixup methods (Meta-FDMixup, base-M3T, TGDM) are better than the non-mixup baselines (m-base, s-base). We explain that the intermediate domain can also help to enhance the diversity of the source domain.

\begin{table*}[]
    \centering
    \begin{tabular}{llllllll}
    \hline
    &Auxiliary Data & Intermediate Domain & Dynamic Mixup & CUB&Cars&Places&Plantae\\
    \hline
    m-base&\checkmark&-&-& 57.65$\pm$0.80&46.03$\pm$0.72&55.70$\pm$0.79&48.25$\pm$0.74\\
    \hline
    base-M3T&\checkmark&\checkmark&-&63.87$\pm$0.26&49.02$\pm$0.24&59.99$\pm$0.26&51.65$\pm$0.25\\
    \hline
    TGDM&\checkmark&\checkmark&\checkmark&\textbf{64.80$\pm$0.26}&\textbf{50.70$\pm$0.24}&\textbf{61.88$\pm$0.26}&\textbf{52.39$\pm$0.25}\\
    \hline
    \end{tabular}
    \caption{Ablation study to verify the effectiveness of each component in TGDM framework. We report the results(\%) on the CUB, Cars, Places, and Plantae benchmarks under the 5-way-1-shot setting.}
    \label{tab:ablation}
\end{table*}

\begin{table*}[]
    \centering
    \begin{tabular}{lllll}
        \hline
        Target Dataset&CUB&Cars&Places&Plantae \\
        \hline
         base-M3T &63.87$\pm$0.26&49.02$\pm$0.24&59.99$\pm$0.26&51.65$\pm$0.25\\
         M3T-fixed &\textbf{66.36$\pm$0.26}&\textbf{51.02$\pm$0.25}&60.80$\pm$0.26&51.71$\pm$0.25\\
         TGDM &64.80$\pm$0.26&50.70$\pm$0.24&\textbf{61.88$\pm$0.26}&\textbf{52.39$\pm$0.25}\\
        \hline
    \end{tabular}
    \caption{The results(\%) of M3T-fixed under 5-way 1-shot setting are reported to verify that our TGDM framework can automatically find an optimal mix ratio for the target datasets.}
    \label{tab:average}
\end{table*}

\subsection{Ablation Study}
We conduct experiment on 5-way 1-shot setting to verify the effectiveness of two components of our framework, which is intermediate domain and dynamic mixup, respectively. The experimental results are shown on Table \ref{tab:ablation}. We design this ablation study in threefold. Firstly, all baseline methods are trained with Mini-ImageNet and auxiliary target data. Secondly, base-M3T and TGDM utilize the intermediate domain to help knowledge transfer from the source domain to the target domain. Concretely, methods ticked in the intermediate domain column use data with source data and auxiliary target data. And base-M3T obtains the mix ratio $\lambda$ by randomly choosing a digit from the Beta(1.0, 1.0) distribution. Thirdly, TGDM uses dynamic mixup guided by target data. Concretely, bi-level training strategy is used to guide the generation of mixed data in the intermediate domain.\par 
From Table \ref{tab:ablation}, we can see that: 1) All the three components are effective, for our framework improves base-M3T by 0.93\%, 1.68\%, 1.89\%, and 0.74\% on CUB, Cars, Places and Plantae, respectively. Moreover, the performance of base-M3T exceeds that of m-base by 6.22\%, 2.99\%, 4.29\%, and 3.4\% on the four target datasets. 2) The m-base is improved by base-M3T significantly, which indicates that the intermediate domain helps to transfer knowledge. Besides, Mixup-3T with tri-task learning can make good use of data of source, target, and intermediate domain. 3) The dynamic mixup component makes further improvements beyond base-M3T, which indicates that our dynamic mixup framework TGDM can find an optimal mix ratio for the target dataset and current model state.

\begin{figure}[!t]
  \centering
   % \vspace{0.1in}
  \includegraphics[width=0.85\linewidth]{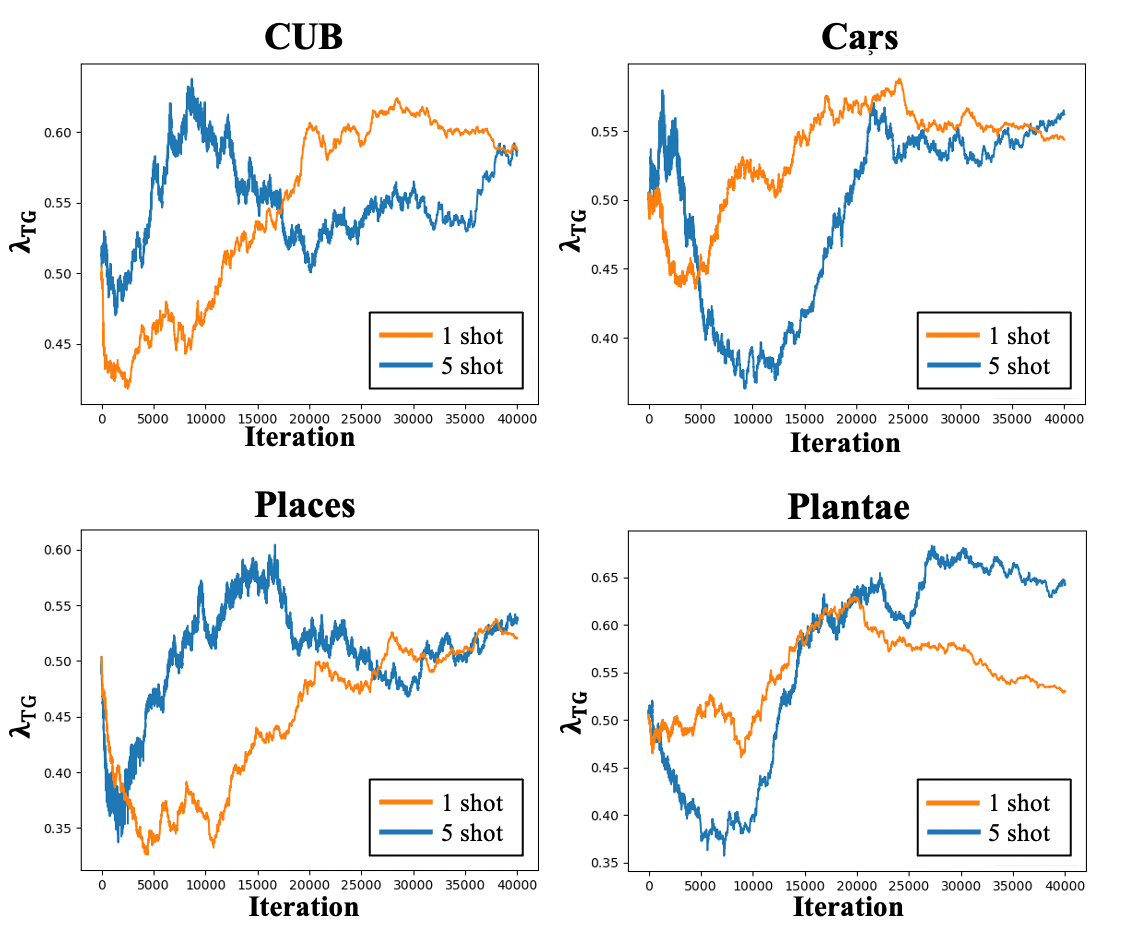}
  \caption{The generated guided mix ratio $\lambda_{TG}$ in each training iteration for CUB, Cars, Places and Plantae datasets.}
  \label{fig:lamda}
  \vspace{-0.15in}
\end{figure}

\begin{figure*}[h]
    \centering
    \includegraphics[width=\textwidth]{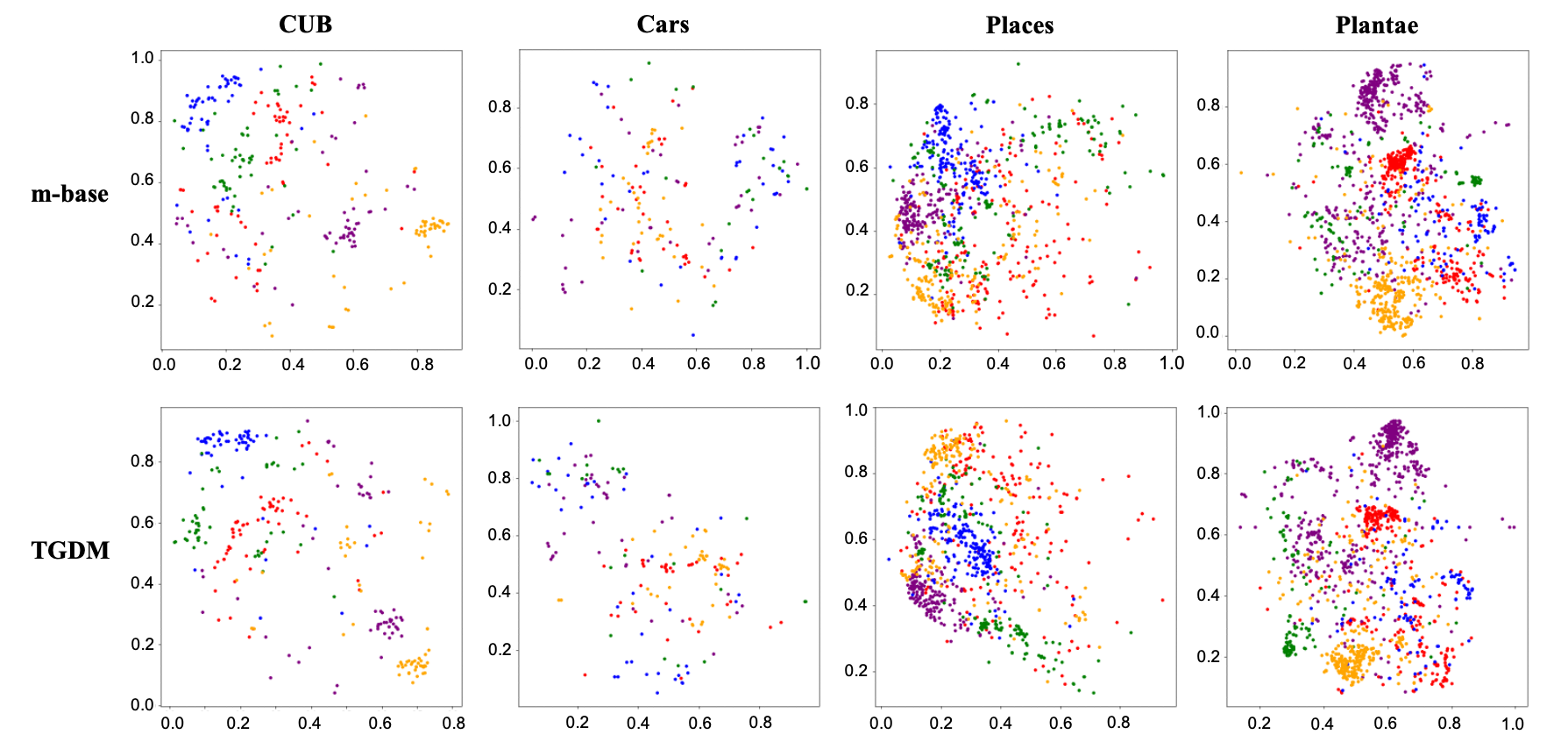}
    \caption{The t-SNE visualization results of our TGDM and m-based under 5-way 5-shot setting. Different color represents different class in an episode, and the episode is randomly chosen from $D_{Tn}$.}
    \label{fig:tsne}
\end{figure*}

\subsection{Analysis}
We first record the target guided generated mix ratio $\lambda_{TG}$ of each epoch for the four target datasets. Intuitively, according to Eq. \ref{eq:mixq}, when the value of $\lambda_{TG}$ approaches to 0, the proportion of target data rises in the mixed data, while when the value of $\lambda_{TG}$ approaches to 1, the proportion of target data drops. 
%\yixin{give intuitive explaination, like xx aaproaches 0, the weight of target domain data becomes higher? } 
From Fig. \ref{fig:lamda}, it can be observed as follows: 1) The mix ratio $\lambda_{TG}$ of each iteration varies from each other, which indicates that the $\lambda_{TG}$ is guided by the target dataset dynamically. Furthermore, it also shows the necessity of learning mix ratio dynamically, instead of manually tuning. For it's impossible to manually tune the mix ratio for every iteration. 2) All the lines in Fig. \ref{fig:lamda} have a very similar trend. Generally, this trend can be summarised as three phases. First, the value of $\lambda_{TG}$ descends and approaches zero in the early iterations. Second, the value rises. Last, the value descends again and converges to around 0.55. There are three major reasons for such phenomenons. 
1) The whole framework is designed in the target guided way, so the value of $\lambda_{TG}$ firstly approaches 0 to raise the proportion of the target data in the first phase. Therefore, the model can achieve good performance on target dataset, which is straightforward. Although for the model trained in cars under 1-shot setting, the line rises at first but descends very quickly around iteration 2500. we explain that it's because the 1-shot learning task is not very stable at the beginning, and the trend drops quickly, which is consistent with our analysis. 
2) In the second phase, the rising of $\lambda_{TG}$ is mainly because of the over-fitting, since the number of auxiliary target data is few. Therefore the proportion of source data increases with $\lambda_{TG}$ rising so that more diversity is added to the training data. 
3) At the last phase, the value of $\lambda_{TG}$ finally converges to a compromise value between avoiding overfitting and achieving good performance in the auxiliary data.
% \begin{itemize}
% \item[1)] The whole framework is designed in the target guided way, so the value of $\lambda_{TG}$ first approaches 0 to raise the proportion of the target data in the first phase. Therefore, the model can achieve good performance on target dataset, which is straightforward. Although for the model trained in cars under 1-shot setting, the line rise at first but descent very quickly around iteration 2500. we explain that it's because the 1-shot learning task is not very stable at the beginning, and the trend drops quickly, which is consistent with our analysis. 
% \item[2)] In the second phase, the rising of $\lambda_{TG}$ is for the reason that number of auxiliary target data is few, which causes and overfitting problem. Therefore the proportion of source data increases with $\lambda_{TG}$ rising so that more diversity is added to training data. 
% \item[3)] At the last phase, the value of $\lambda_{TG}$ finally converges to a compromise value between overfitting and performanfce of the target domain.
% \end{itemize}

In order to further illustrate that our TGDM framework can find the optimal mix ratio for the target dataset. We average $\lambda_{TG}$ over all the iterations as $\lambda_{avg}$ and then test the performance when the mix ratio is fixed to $\lambda_{avg}$. For simplicity, we define this method as M3T-fixed. The average $\lambda_{TG}$ value for CUB, Cars, Places, and Plantae are 0.5499, 0.4909, 0.5057 and 0.5632, respectively. The results of M3T-fixed under 5-way 1-shot setting are recorded in Table \ref{tab:average}. From  Table \ref{tab:average}, it can be seen that the performance of M3t-fixed is much higher than base-M3T in all datasets. And at some cases, the performances of M3T-mixed are even higher than the TGDM framework. We explain that this is because M3T-fixed utilizes the optimal mix ratio found by TGDM at the beginning of the training process. These results shows that the simple average operation can still find an effective mix ratio, which further indicates that our TGDM framework can automatically find an optimal mix ratio for the target dataset instead of a manually searching process.
\subsection{Visualization Result}
To intuitively show how our method improves knowledge transfer, we visualize the features extracted by Mixup-3T through t-SNE in Fig. \ref{fig:tsne}. For visualization, we randomly sample five classes from $D_{Tn}$. Then features are obtained by the feature extractor (ResNet10) in Mixup-3T. After that, all the features are projected into 2-D space. For comparison, we also conduct visualization for m-base model.
From Fig. \ref{fig:tsne}, it can be observed that compared to m-base model, the feature distributions of our method are more distinct and the inter-class distance is smaller. As a result, it makes the classifier easy to recognize the image and thus achieves better classification performance.
%\yixin{cannot see the analysis results from the fig. I would suggest to remove this section and add some case study.}

\section{Conclusion}
In this paper, we proposed the TGDM framework that tackles the CD-FSL learning by introducing a target guided intermediate domain for knowledge transfer. The proposed TGDM framework consists of two modules: Mixup-3T for classification and DRGN for dynamic mixup ratio generation. A tri-tasks learning mechanism is adopted in Mixup-3T in order to transfer knowledge with the help of the intermediate domain. Besides, bi-level meta-learning strategy is utilized together with DRGN to produce a target guided mix ratio and intermediate domain data, which are optimal for the current model and the target domain. Extensive experimental results on several benchmark datasets verify the effectiveness of our method.

\section{Acknowledgement}
This work was supported in part by National Natural Science Foundation of China Project (No. 62072116), Shanghai Pujiang Program (No. 20PJ1401900) and Shanghai Science and Technology Program (No. 21JC1400600).

\bibliographystyle{ACM-Reference-Format}
\balance
\bibliography{main}

\end{document}